\documentclass[runningheads,hidelinks]{llncs}
\usepackage{graphicx}

\usepackage{tikz}
\usepackage{comment}
\usepackage{amsmath,amssymb}
\usepackage{color}
\usepackage{hyperref}
\usepackage{array}

\usepackage[accsupp]{axessibility}

\begin{document}

\pagestyle{headings}
\mainmatter

\title{An extensible point-based method for data chart value detection}
\author{Carlos Soto and Shinjae Yoo}
\institute{Brookhaven National Laboratory, Upton NY 11973, USA\\
\email{csoto@bnl.gov}
}

\maketitle

\begin{abstract}
We present an extensible method for identifying semantic points to reverse
engineer (i.e. extract the values of) data charts, particularly
those in scientific articles. Our method uses a point proposal network
(akin to region proposal networks for object detection) to directly predict
the position of points of interest in a chart, and it is readily
extensible to multiple chart types and chart elements. We focus on complex bar charts in the
scientific literature, on which our model is able to detect salient points with an accuracy of 0.8705 F1 (@1.5-cell max
deviation); it achieves 0.9810 F1 on synthetically-generated charts similar to those used in prior works.
We also explore training exclusively on synthetic data with novel augmentations,
reaching surprisingly competent performance in this way (0.6621 F1) on real charts with widely varying
appearance, and we further demonstrate our unchanged method applied directly to synthetic pie
charts (0.8343 F1). Datasets, trained models, and evaluation code are available at
\url{https://github.com/BNLNLP/PPN_model}.

\end{abstract}

\section{Introduction}
\label{sec:intro}

A long-standing challenge in document information mining is the general
inaccessibility to machine processing of data contained in embedded images.
Although established computer vision techniques (e.g. OCR) and more recent
ML-based ones -- like image classification/captioning, object detection, and
similarity scores -- have alleviated this limitation in some contexts
(e.g. \cite{algorri2007automatic,mondal2020iiit,rajasekharan2010image,jain2019multimodal}),
identification and extraction of structured image data
remains a challenge. This is particular the case for scientific literature
mining, in which much of the new knowledge presented in each article is in
the form of data-filled charts.

\begin{figure}
  \centering
  \includegraphics[width=1.0\linewidth]{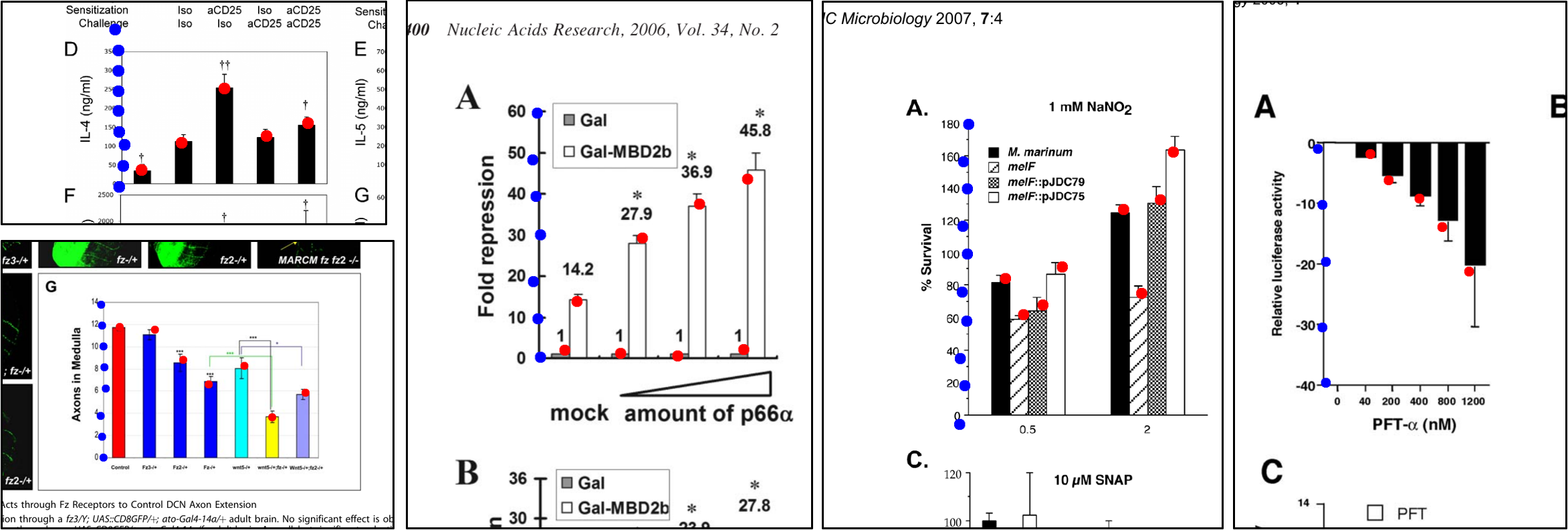}
  \caption{Sample detection results of data points in scientific bar charts. Images intentionally over-cropped, showing robustness to spurious artifacts. Charts from \cite{rocha2003p53,brackertz2006p66alpha,srahna2006signaling,subbian2007mycobacterium,lewkowich2005cd4}.}
  \label{fig:sample-results}
\end{figure}

There is of course immense scientific and practical value in collating knowledge from vast and disparate document sources: enabling broad search, powerful inference, new analyses, and a myriad other capabilities that are out of reach when considering documents individually or in small sets as human readers are limited to doing. Document mining has already impacted the state of researcher access to knowledge found in the scientific literature: common tools like keyword search and reference crawling may now be supplemented by text mining of abstracts (or even full articles); scientific text corpora are designed and published to enable such mining (e.g. \cite{roberts2001pubmed,wang2020cord,kononova2019text}); and ML models such as SciBERT \cite{beltagy2019scibert} and BioBERT \cite{lee2020biobert} have exploited such datasets to give researchers new capabilities leveraging natural language processing techniques. But thus far, automated and scalable information extraction has been limited mainly to text, despite the wealth of information found in charts and graphs.

The value of information found in data charts is certainly not limited to the scientific literature; business, government, educational, historic, and many other types of document sources could benefit from elucidation of their graphically-embedded contents. But \textbf{charts in published research are of particular interest to us because they ostensibly form a part of the scientific record}, yet their format impedes access to the information they contain. Presenting data graphically has obvious benefits, \textbf{but if the underlying data is not also accessible, reproducibility and comparison become difficult}. There are increasing efforts to encourage publication of research data (including the raw data that comprise chart contents), but this practice is not yet the norm. And there is certainly no expectation that the figures in millions of published articles will be retroactively annotated with their source data values. Even attempting this for small numbers of charts is a daunting and tedious challenge.

It would therefore be highly desirable to be able to automatically reverse engineer chart contents. Manual annotation tools such as WebPlotDigitizer \cite{marin2017webplotdigitizer} and GraphReader \cite{graphreader} may be useful for single or small numbers of graphs, but these of course do not scale beyond such use cases -- and the manual annotations they rely on mean they cannot be well adapted to automation. This is further complicated by the tremendous variety of graph types, styles, production tools, overlays, and other visualization effects that authors and publishers employ.

\textbf{In this paper, we address what we believe is the most pressing part of the automated data extraction problem: accurate value detection in complex charts; we focus our efforts on scientific bar charts. However, the method we propose is designed to be readily extensible to other chart types (as we demonstrate with pie charts) and to fit naturally in a full data extraction pipeline}. The approach we take is point-based, and focuses on the detection of salient data points in chart images. Figure \ref{fig:sample-results} shows sample results of our value detection method on a number of scientific bar charts, demonstrating robustness to varying styles and spurious artifacts. Note that this work does \textit{not} include the detection of text labels nor the alignment of these with detected data points to complete the value extraction process.

Source code and trained models are available (link in abstract), and we also contribute several datasets, including 500 manually-annotated bar charts from scientific articles, which we used for our model fine-tuning and evaluation; these are significantly more varied in appearance and the presence of artifacts than previous synthetic and web-sourced datasets (e.g. those used in \cite{liu2019data,zhou2021reverse,ma2021towards,davila2019icdar,davila2021icpr}). Additionally: 1) we show that our approach significantly outperforms a capable recent method \cite{zhou2021reverse}, 2) we assess the use of data augmentation to better exploit synthetic data, and 3) we demonstrate that our method can be directly and successfully applied to other chart types -- pie charts in particular.

\section{Related Work}
\label{sec:relatedwork}

Ours is not the first work to investigate information extraction from data charts. ReVision \cite{savva2011revision} and ChartSense \cite{jung2017chartsense}, for example, demonstrated accurate classification of different types of charts images collected from the web, and ReVision further used a set of hand-crafted features and rules to extract salient marks and use these to infer chart values. ChartSense, on the other hand, is an interactive extraction tool (similar to \cite{marin2017webplotdigitizer} and \cite{graphreader}), but includes OCR for automatic text element labeling. Other works, such as REV \cite{poco2017reverse}, ChartReader \cite{rane2021chartreader}, and that by Al-Zaidy and Giles \cite{al2015automatic}, have taken a combined approach, using rule/heuristic-based element extraction supplemented by OCR for text elements (ChartReader also uses ML-based chart classification). The problem is of course that rules and hand-crafted features are brittle and must be reproduced and tuned for each new chart type -- or even for different styles and contexts.

Deep learning approaches to chart value extraction have primarily leveraged object detection methods. Liu et al. \cite{liu2019data} and Zhao et al. \cite{zhou2021reverse}, for example, apply Faster R-CNN \cite{ren2015faster} to detect chart elements; Scatteract \cite{cliche2017scatteract} uses an object detector called ReInspect \cite{stewart2016end}; and Ma et al. \cite{ma2021towards} use Cascade R-CNN \cite{cai2018cascade} for bounding box detection, as well as a segmentation-based point detector.
With the notable exception of \cite{rane2021chartreader} (which reported detection accuracies between 22.98\% and 42.10\% for data values in bar charts), prior works have generally been limited to testing and demonstration on synthetic or otherwise very clean chart images.

In contract to these and similar methods (see Davila et al. \cite{davila2020chart} for a recent survey), we opt to exclusively use point-based value detection, as we believe it is more readily adaptable to detecting the salient elements in different chart types, is easier to annotate, can be readily extended to additional element types (e.g. error bars and stacked charts), and better supports integration with additional element labeling (e.g. OCR). Such point-based detection has not been common in chart value extraction, but has seen used for other purposes, such as radioastronomy and particle imaging \cite{Domine2021point,tilley2020point}. In this work, we target complex scientific charts, and so we particularly aim to reduce prior works' assumptions about image simplicity, regularity of style and appearance, and presence of artifacts.

\section{Preface: Isolating Bar Charts in Documents}
\label{sec:detectingcharts}

Visual value extraction of document-embedded charts \textit{could} be attempted with whole-page images; however in this work we separate the problem of chart detection from value detection, and focus on the latter for several reasons: 1) There are already strong vision methods which can be adapted for detecting charts of a particular class or recognizing charts among all extracted document figures -- as well as document layout analysis methods and datasets for segmenting document pages into their several constituent regions \cite{zhong2019publaynet,soto2019visual,oliveira2018dhsegment,li2020docbank}. 2) In many cases, images can be extracted intact from documents (e.g. those in HTML/XML formats as in PubMed Central, and supplementary material or archived sources as found in arXiv), further simplifying that problem to a classification one. And finally, 3) because image resolution must be preserved at a quality that enables useful value extraction, doing so with whole-document images would require significant (and largely wasted) computational and memory expense.

Although detection of bar chart figures is not the focus of this work, we did seek to verify for ourselves the performance and effects of applying existing methods to this task. So in addition to the set of bar charts we value-annotated, we also labeled a much larger dataset of isolated bar chart bounding boxes (Sec. \ref{subsec:chartdetdataset}), and used it to train a standard YOLOv3 \cite{redmon2018yolov3,eriklindernoren} object detection model. Our testing showed that the model (pre-trained on COCO \cite{lin2015microsoft}) could be readily fine-tuned to detect bar charts with a mAP (mean average precision) of 95.2\% at 0.5 IoU (intersection-over-union), though detection performance drops significantly at higher precision thresholds (88.3\% @0.75 IoU, 18.1\% @0.9 IoU).

As higher IoU corresponds to better-fitting bounding boxes, our tentative conclusion is that standard object detection models can accurately detect and localize bar charts, but may be less capable of precisely determining the chart bounding boxes (\textit{Note that we made no significant modifications to the YOLOv3 model in our testing; there is likely plenty of room for improvement}). A value extraction method such as ours (Sec. \ref{sec:ppn}), if used in a pipeline with such chart detection methods, would therefore need to contend with spurious artifacts from poor cropping of detected charts. One practical way to address this is to intentionally over-crop the detected charts before sending them to the value detection model, and training that model using data designed with this expectation. We followed precisely this approach, as described in Sec. \ref{subsec:synthdata} - \ref{subsec:realaug}.

\section{A Point Proposal Network for Chart Value Detection}
\label{sec:ppn}

\begin{figure}
  \centering
  \includegraphics[width=1.0\textwidth]{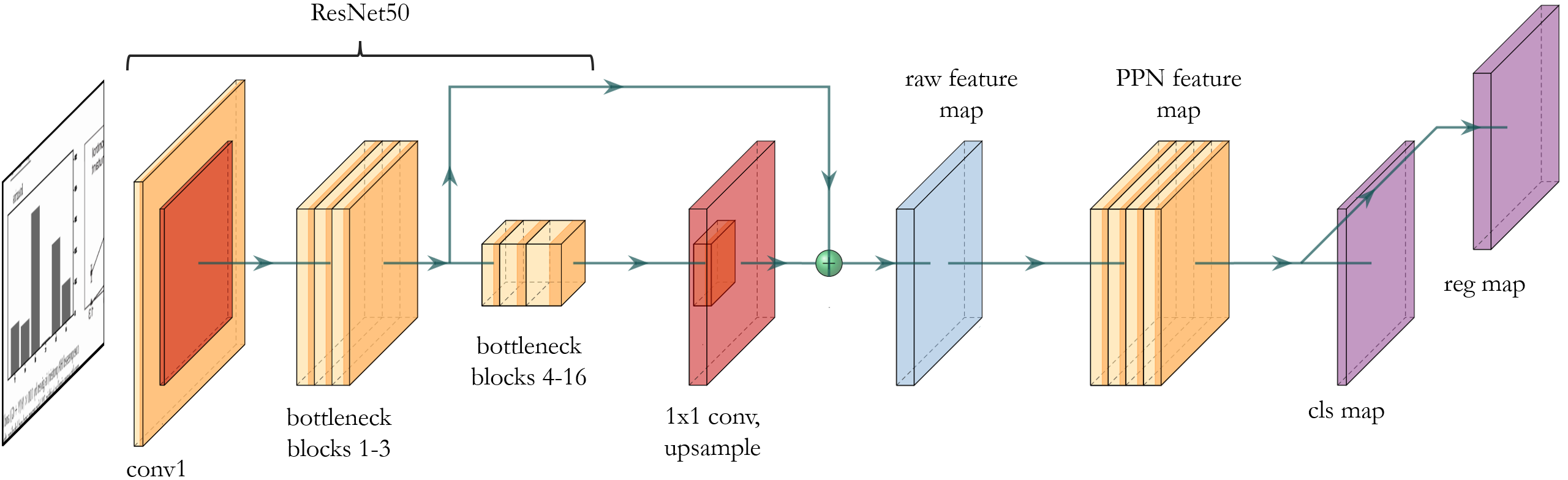}
  \caption{Architecture summary for our Point Proposal Network (PPN). Features are extracted using a ResNet50 backbone; the main PPN constitutes a 4-layer CNN with a classification and regression head, producing class and point-tuning regression maps.
  }
  \label{fig:arch}
\end{figure}

We designed a convolutional neural network model which predicts a set of points in an image, and classifies them into a set of defined labels -- directly analogous to object detection, but restricting the problem to predicting zero-size points rather than bounding boxes. We believe this is better suited to the problem of value extraction in different chart and graph types, whereas object detection models may not provide the same flexibility. For example, an object detection model may be readily adapted to detect elements in bar or pie charts (e.g. \cite{liu2019data}), but it is not clear how it might be applied to extracting values from line graphs, dense histograms, stacked charts, etc. Furthermore, a point-based approach supports additional granularity, such as identifying error bars. To support bar charts of arbitrary appearance, we detect two element types: the peaks of each bar, and the labeled tick marks along the value axis (but not the text labels themselves). The chart origin is treated as a tick mark special case.

Our model (Figure \ref{fig:arch}) uses a ResNet50 backbone \cite{he2016deep} for feature extraction at an input image resolution of {224x224}, with two feature maps extracted after the 3rd (56x56, 256-channel) and final bottleneck blocks (7x7, 2048-channel). A 1x1 convolutional layer is used to reduce the channel depth of the second feature map to 256, and it is then up-sampled to 56x56 via bilinear interpolation and added to the first feature map. This approach was taken to preserve both deep semantic features and fine detail -- necessary for data value extraction. The 56x56 feature map resolution was chosen for its natural fit within the existing ResNet50 model, and for consistency with the maximum density of data points expected in scientific bar charts (\textit{denser charts may be encountered or produced; we settled on a map size that covers a significant majority of useful cases}).

\begin{figure}
  \centering
  \includegraphics[width=1.0\linewidth]{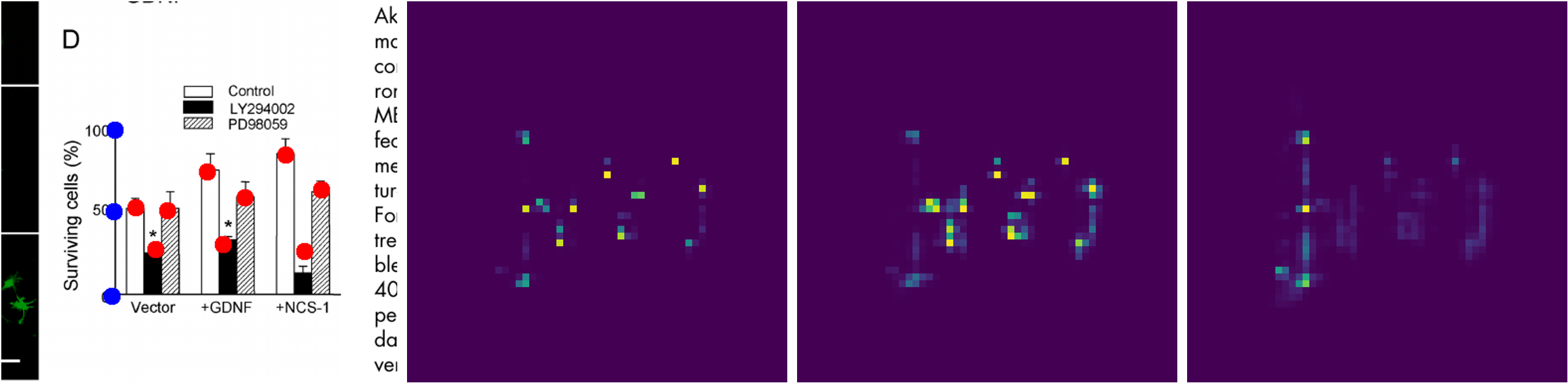}
  \caption{An example of the class maps produced by our PPN. Left-to-right: input image superimposed with our model's bar and tick point predictions, the background point map (negated), bar class map, and tick class map. The source chart image is intentionally over-cropped. (Sample chart is from \cite{nakamura2006novel})}
  \label{fig:maps}
\end{figure}

The feature map passes through a further 4-layer convolutional Point Proposal Network (PPN) with two per-pixel prediction heads: a classifier and a fine-tuning regression head. This is akin to region proposal networks used in e.g. Faster R-CNN \cite{ren2015faster}, except that the PPN can make point predictions directly, without any need for ROI (region-of-interest) cropping to classify (and further refine) bounding boxes. The classification head predicts one of three classes: background, bar, and tick; the regression head predicts the point offset. Figure \ref{fig:maps} shows examples of the class maps produced for a sample chart.

\subsection{Predicting Data Points}
\label{subsec:nms}
After the PPN model produces 56x56 classification and regression maps, these are converted to a list of predicted points for each bar and tick. This prediction is done by first masking the class map at a threshold over the background class (in practice, we found values between 0.9 and 0.99 work well), and then selecting pixels that were classified as bars or ticks with scores greater than some threshold (we found 0.75 works well). The corresponding values in the regression map are used to tune the position of each detected point, and the resulting points are passed through a non-maximum suppression (NMS) function to remove duplicate detections. Our NMS implementation sorts points in each class by confidence score and iteratively removes low-scoring points within a threshold radius (we used a default threshold of 1.5).

For the bar class, we made the NMS threshold ellipse anisotropic, giving it a 5x greater reach in the horizontal direction, as bar widths can vary significantly from chart to chart. The final list of bar and tick point predictions can then be visualized (as in Figures \ref{fig:sample-results},\ref{fig:maps}), used together with a text label detector to extract numerical values (not covered in this work, see Sec. \ref{subsec:limits}), or compared to sets of ground truth points during training, as discussed in the following section.

\subsection{Loss Function and Training}
\label{subsec:lossfunc}
During training, the model has access to the input image and to corresponding ground truth labels for the lists of bar and tick points, and a class and regression map generated from these lists. The PPN model, together with the map-to-point conversion and NMS described in Sec. \ref{subsec:nms}, generates predictions for each of these, so the loss function consists of terms for each target as well. In theory, we could derive a loss from just the lists of points, but we suspected that this alone would make for a poor training signal -- especially at earlier epochs -- and confirmed as much, finding that such an approach fails to train effectively. Our loss function is then
\begin{equation}\label{total_loss}
   L = \lambda_1 L_{cls} + \lambda_2 L_{reg} +  \lambda_3 L_{pts} +  \lambda_4 L_{align}.
\end{equation}
The class map loss $L_{cls}$ is a regular cross-entropy loss, and the regression map loss $L_{reg}$ uses a masked mean squared error:
\begin{equation}\label{reg_loss}
   L_{reg} = \sum_{i,j}(\hat{r}_{i,j} - r_{i,j})^2, \mathrm{if}\; \hat{c}_{i,j} \ne background,
\end{equation}
where $\hat{r}_{i,j}$,$r_{i,j}$, and $\hat{c}_{i,j}$ are, respectively: the predicted regression value, ground truth regression value, and predicted class value at map position $i,j$. The points list loss is
\begin{equation}\label{pts_loss}
   L_{pts} = \sum_{p}dist(\hat{p}, p), \mathrm{if}\; dist(\hat{p}, p) < T_{det},
\end{equation}
where $\hat{p}$ and $p$ are predicted and ground truth within-class points, $dist$ is Euclidean distance, and $T_{det}$ is a max detection distance threshold (default=1.5).

We followed a training schedule in which $\lambda_{3}$ starts at zero and is ``turned on" only when the training loss begins to plateau. At that point it is weighted to $1/10$ relative to $\lambda_{1,2}$, due to scale differences. We did not find training to be particularly sensitive to these other term weights. When training had again plateaued, we added a fourth term to the loss function to promote horizontal alignment of the predicted tick marks:
\begin{equation}\label{align_loss}
   \lambda_4 L_{align} = \sum_{tp}abs(\widehat{tp}_x - mean(tp_x))
\end{equation}
This tick alignment loss is an example of a chart-type-specific loss term; we expect analogues can be designed for other chart types as well. We set $\lambda_4 = 1/1000$ relative to $\lambda_{1,2}$, again due to scale differences. We trained our model with regular stochastic gradient descent, using a learning rate of 1e-3, momentum = 0.9, and a batch size of 64. Training took about 3-4 minutes per epoch (for 5000-image datasets) on two NVIDIA 1080Ti GPUs.

\section{Datasets}
\label{sec:datasets}

We annotated and generated five different datasets, described here. The first is for bar chart detection, on which we evaluated baseline performance and detection qualities of a standard object detection model (see Sec. \ref{sec:detectingcharts}). The following three datasets are for the main problem this work addresses: data point detection in bar charts to support automated value extraction. The last dataset is for value detection in pie charts. All datasets and scripts are available at \url{https://github.com/BNLNLP/PPN_model}.

\begin{table}
\centering
\caption{
  \label{tab:datasets}
  Summary of dataset types, uses and statistics.
}
\begin{tabular}{lllll}
\hline
\textbf{Dataset} & \textbf{Type} & \textbf{Use} & \textbf{Images} & \textbf{Labels} \\
\hline
5.1 & Manual & Bar chart detection &  1592 & 3548 \\
5.2 & Synthetic & Value detection (bar charts) & 5000 & 65618 \\
5.3 & Manual & Value detection (bar charts) & 500 & 7109 \\
5.4 & Augmented & Value detection (bar charts) & 5000 & 65618 \\
5.5 & Synthetic & Value detection (pie charts) & 5000 & 32718 \\
\hline
\end{tabular}
\end{table}

\subsection{Bar chart detection}
\label{subsec:chartdetdataset}
We collected 673 article PDFs from the PubMed Central Open Access Subset \cite{pmcoa}, and rendered each page at 72dpi (using the pdftoppm utility from the standard poppler-utils package), resulting in 1592 document page images with a typical resolution of 612x792. We then manually labeled bounding boxes for each bar chart in each page (using LabelImg \cite{tzuta2015github}), whether they were standalone figures or were embedded in part of another figure. The selection criteria we followed excluded similar but distinct chart types such as histograms, and we avoided including figures that contained multiple types of overlapping charts (e.g. a line graph superimposed on a bar chart) or charts with multiple data points per category (e.g. stacked charts or waterfall/floating charts). We did, however, include bar charts in any orientation (horizontal, vertical, upside-down), at any size, in any color, and with or without: error bars, 3D effects, legends, category/data labels, or embedded titles/captions/comments or other artifacts. The only strict criteria was that each chart must include a labeled value axis.

In total, we annotated bounding boxes for 3548 scientific bar charts, of which only 144 were horizontal (a 24:1 ratio). This observation informed our decision to restrict our other datasets to vertical bar charts only, as there would be disproportionately few horizontal ones to sample from, and we expect our method and results to generalize without issue to those, with only minor adaptation.

\subsection{Synthetic bar charts for value detection}
\label{subsec:synthdata}
We used the Matplotlib library in Python to programmatically generate large numbers of bar charts with randomly varying numbers of bars (2-10), bar widths (0.1-0.9), colors (123 choices from the CSS4\_COLORS space, excluding those deemed too light), and values. We restricted our generation to plain bar charts: no stacked charts, no grouped charts, and no error bars. This is due to this dataset only being used for pre-training. Nonetheless, these synthetic charts are similar in complexity to those used in previous works (e.g. \cite{liu2019data,zhou2021reverse,ma2021towards}).

We generated 5000 charts (80:20 training:val split), and augmented these simply by applying random padding (up to 50\% per dimension) and cropping (taking care to preserve the full chart within cropping bounds). Bar and tick mark annotations were extracted programmatically during chart generation. We also generated a separate test set of 1000 charts in the same manner. 

\subsection{Manually annotated real charts values}
\label{subsec:m246}
We randomly sampled 500 vertical-only bar charts from dataset \ref{subsec:chartdetdataset} and re-rendered the PDF pages containing them at 300dpi. We also used the bounding box annotations from \ref{subsec:chartdetdataset} to produce crops of the bar charts at high resolution, with crops padded by 50\% to intentionally incorporate parts of surrounding text, nearby figures and other artifacts. This is to accommodate the expectation that automatic bar chart extraction may yield imperfect crops (see Sec. \ref{sec:detectingcharts}), so a value detection method should be robust to such artifacts.

We then used the online tool \href{https://www.makesense.ai/}{makesense.ai} to manually label the bar, tick, and origin points in each chart as precisely as practical (temporary image recoloring and contrast adjustments made this process easier). We estimate that our manual annotations are generally within 3 pixels distance of their true value (median image size 663x475, range approximately $\pm$3X). We converted these annotations to the simple label format we designed for \ref{subsec:synthdata}.

\subsection{Real-augmented synthetic charts}
\label{subsec:realaug}
As it is much quicker to annotate whole chart bounding boxes than individual data points in those charts (compare image and label numbers in datasets \ref{subsec:chartdetdataset} and \ref{subsec:m246} in Table \ref{tab:datasets}) we sought to exploit annotations of the former type to augment our synthetically generated bar charts (dataset \ref{subsec:synthdata}). A significant amount of the variance in a published scientific chart is found in its immediate surroundings (i.e. not only in its interior contents), so we decided to use these surroundings to improve the local feature learning of our convolutional PPN model.

We took random crops from the regions immediately adjacent to the bar chart bounding boxes in dataset \ref{subsec:chartdetdataset} (up to 50\% beyond the chart bounds) and pasted these to the padded surroundings of our synthetic charts. Before doing so, we randomly applied morphological erosion (up to 4 pixels) and dilation (up to 1 pixel) to the synthetic charts in order to diversify their line thicknesses. Finally, we randomly cropped the chart images (keeping the full chart in frame) and rescaled along each dimension by up to 50\%.

\subsection{Synthetic pie charts}
\label{subsec:synthpie}
To demonstrate the extensibility of our method to other chart types, we also generated a dataset of synthetic pie charts with Matplotlib. For these, we produced two classes of point annotations: the boundaries between pie wedges, and the pie chart centroid. From these points, the size/values of the pie chart elements (and thus the chart's original data) can be readily determined.

\section{Experiments}
\label{sec:experiments}

We trained and evaluated our model on synthetic data \ref{subsec:synthdata}, manually annotated real data \ref{subsec:m246}, and real-augmented synthetic data \ref{subsec:realaug}. Our primary evaluation metric for bar charts is the accuracy of data point predictions for bar peaks and value-axis tick marks. We also compare our method to \cite{zhou2021reverse} and demonstrate easy extensibility of our approach to other chart types, particularly pie charts.

\subsection{Metrics}
We evaluated our models' data point predictions using F1 at a fixed precision threshold measured in units of the PPN class map. As our model only predicts a maximum of one data point per map cell, we chose a precision threshold of 1.5 cell widths to determine prediction recall and precision. The classification map has a 56x56 resolution, so this corresponds to a maximum error of 1.5 / 56 = 2.68\% (a stricter precision criterion than used in e.g. \cite{zhou2021reverse,rane2021chartreader}). At this threshold, we can determine how accurately the model detects the location of data points of interest, without setting an overly strict precision requirement. Empirically, we found this threshold to be the coarsest whose prediction results looked `close enough' (all sample results in Figures \ref{fig:sample-results},\ref{fig:maps}, and \ref{fig:aug-samples} were produced at this threshold), but we also evaluated performance at other thresholds (see Figure \ref{fig:plots}D).

We chose to set the detection threshold with respect to the fixed-size feature map, rather than in image space, in order to be representative of our chart value extraction use case: error should be proportional to the size of the chart. For the same reason, F1 is evaluated \textit{per image} then averaged over the test set. If we instead measured F1 over all target points in an image set, results would appear artificially better ($\sim$5-10\% uplift) because good performance on charts with many points outweighs poor performance on charts with few points.

\subsection{Results}

\begin{table}
\centering
\caption{
\label{tab:results}
Experimental results for our model, trained and evaluated on our datasets (see Table \ref{tab:datasets}). All results on test data unseen during training or hyperparameter optimization, measured at precision thresh=1.5. Real scientific charts in bold.
}
\begin{tabular}{l m{5mm} l m{5mm} l}
\hline
\textbf{Trained on} & & \textbf{Evaluated on} & & \textbf{F1} \\
\hline
5.2 (synthetic bar charts) & & 5.2 & & 0.9810 \\
 & & \textbf{5.3} & & \textbf{0.5549} \\
\textbf{5.3 (real bar charts)} & & \textbf{5.3} & & \textbf{0.8705} \\
5.4 (augmented synthetic bar charts) & & 5.4 & & 0.9307 \\
 & & \textbf{5.3} & & \textbf{0.6621} \\
\hline
\end{tabular}
\end{table}

\begin{figure}
  \centering
  \includegraphics[width=0.9\linewidth]{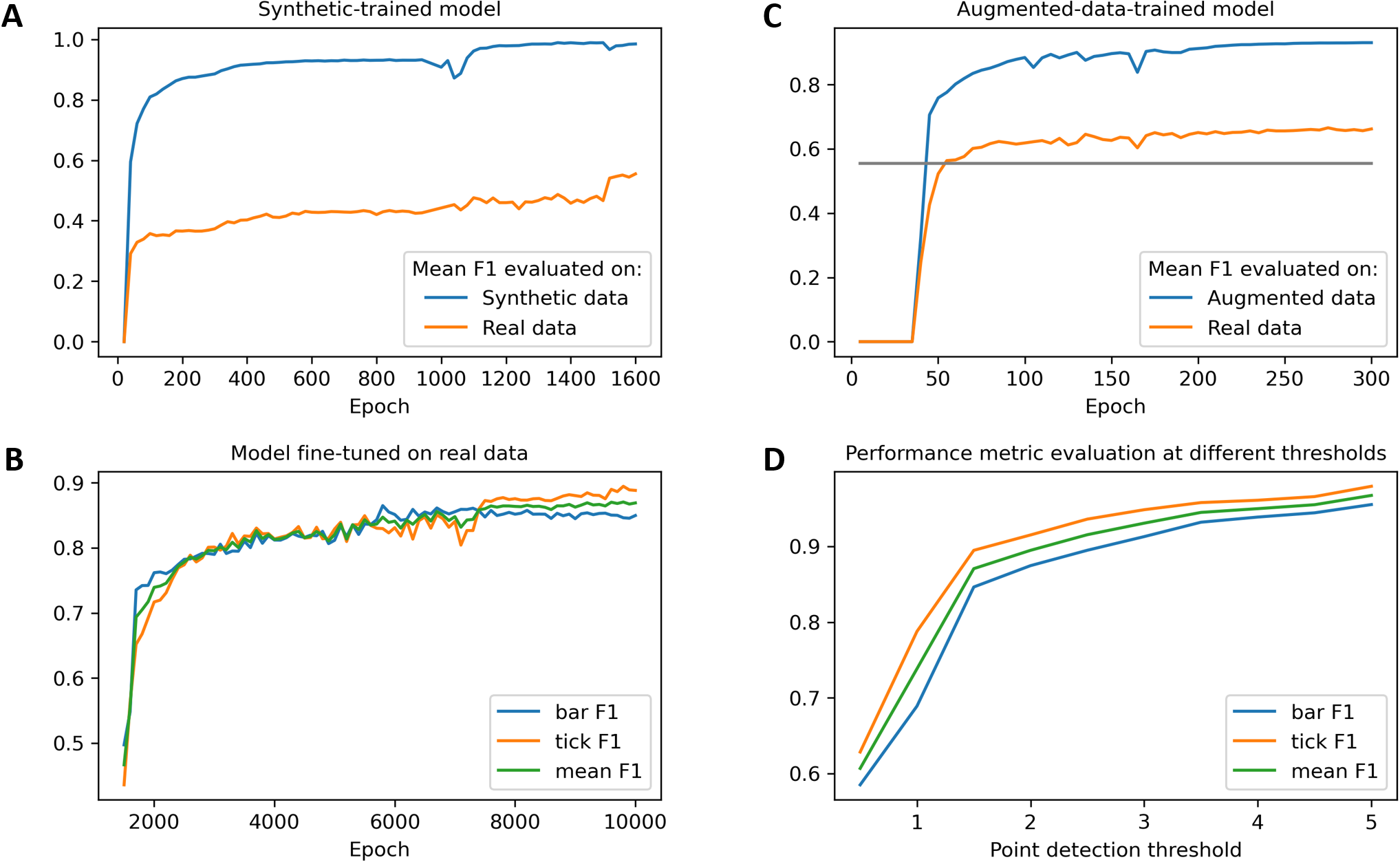}
  \caption{
    \textbf{A}: Synthetic-trained model performance on synthetic and manually-annotated real data.
    \textbf{B}: Model tuned on manually-annotated real data.
    \textbf{C}: Model trained on real-augmented synthetic dataset. The horizontal line is the peak performance (on real data) of model trained on regular synthetic data.
    \textbf{D}: Impact of different precision thresholds on detection performance (same model as B). Thresh=1.5 used for A-C.}
  \label{fig:plots}
\end{figure}

We first pre-trained our model on synthetic data only. As described in \ref{subsec:synthdata}, these synthetic charts were quite simplistic and so had a pronounced `artificial' look to them (similar to those seen in prior works) that is not necessarily representative of published scientific charts. As shown in Table \ref{tab:results} and Figure \ref{fig:plots}A, the pre-trained model achieves strong performance on synthetic data (0.9810 F1), but is predictably less capable when evaluated on real data (0.5549 F1). Note the effects of our training schedule (Sec. \ref{subsec:lossfunc}): $L_{pts}$ is added at epoch 1000 and has a pronounced effect on prediction in synthetic data; $L_{align}$ is added at epoch 1500 and has a more significant impact on real-chart performance.

We next continued training by fine-tuned on our manually annotated dataset. Figure \ref{fig:plots}B shows evaluated performance results on a test-only holdout set of 54 charts. We again followed a multi-step training schedule: $L_{pts}$ was enabled after epoch 5000 and $L_{align}$ was added after epoch 7000 (note that detection performance on tick marks improves markedly after that point). Our model ultimately achieved a mean F1 of 0.8705 on real test data (see Figure \ref{fig:sample-results}).

\begin{figure}
  \centering
  \includegraphics[width=1.0\linewidth]{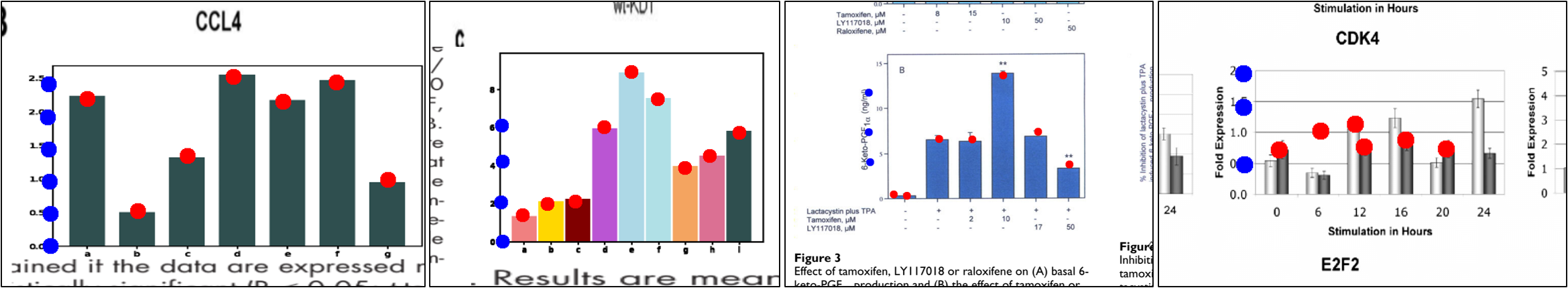}
  \caption{Sample results for model trained only on augmented synthetic data. Left two: synthetic test samples, showing real-document crops pasted at edges. Right two: real charts from scientific articles (from \cite{levine2004tamoxifen,walker2005mnt}). Note the increased detection errors.}
  \label{fig:aug-samples}
\end{figure}

\begin{figure}
  \centering
  \includegraphics[width=1.0\linewidth]{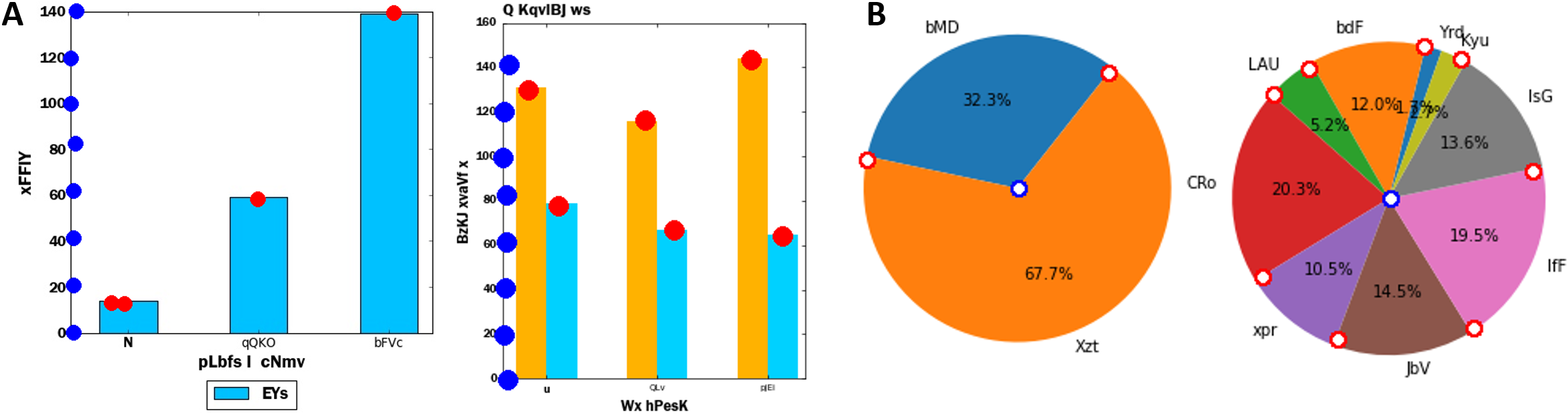}
  \caption{
    \textbf{A}. Sample results of our model on charts from \cite{zhou2021reverse,zhou2021github}.
    Note the similarity to our synthetic samples.
    \textbf{B}. Sample results of our unmodified model trained on synthetically generated pie charts,
    demonstrating direct method extensibility.
  }
  \label{fig:zhou-pies}
\end{figure}

We also assessed the ability of our model to learn value detection for real scientific bar charts by training only on synthetic data that was augmented with fragments of unannotated real chart images (dataset \ref{subsec:realaug}). As shown in Figure \ref{fig:plots}C, training the same model on this augmented data allows it to significantly outperform a model trained on regular synthetic data (even with 5x less training). In this case, $L_{pts}$ was enabled after epoch 100 and $L_{align}$ after epoch 200. This model achieved 0.6621 F1 on real data. Figure \ref{fig:aug-samples} shows samples of this model's detection performance on real-augmented synthetic data and real data.

\begin{table}
\centering
\caption{
  \label{tab:zhou}
  Comparison of our model to  \cite{zhou2021reverse}. All results on \cite{zhou2021github} test set. Our evaluation criteria is 1.5 cell (2.7\% deviation); the criteria used in \cite{zhou2021reverse} is 5\% deviation (2.8 cell),
}
\begin{tabular}{l m{3mm} l m{3mm} l}
\hline
& & \multicolumn{3}{c}{Performance (F1) per eval criteria}\\
\textbf{Model} & & \textbf{Ours (1.5 cell)} & & \textbf{\cite{zhou2021reverse} (5\% deviation)} \\
\hline
\cite{zhou2021reverse} & & & & 0.91 \\
 Ours (unmodified) & & 0.8959 & & \textbf{0.9346} \\
 Ours (tuned on \cite{zhou2021github} training set) & & \textbf{0.9718} & & \textbf{0.9727} \\
\hline
\end{tabular}
\end{table}

We compare our method to a recent work by 
Zhou et al. \cite{zhou2021reverse} which also aimed to release their 
synthetic and real data. Unfortunately, in their GitHub \cite{zhou2021github}, they report they 
have lost their experimental data (presumably both 
types), but reproduced the synthetic data from 
the same procedure as in the paper (on which their method 
achieved 0.91 F1 score at a deviation criterion of 5\%.
We note that the visual complexity of the charts in this dataset
is similar to those in other works (see Sec. \ref{sec:relatedwork}).
We evaluated our method against Zhou’s test dataset, at the 
same 5\% deviation criteria (equivalent to 2.8 cell widths on 
our 56x56 classification grid), as well as at our own, stricter default 
precision threshold of 1.5 cell widths (2.7\% deviation).
Quantitative results are in Table \ref{tab:zhou}, showing that
both our unmodified method (i.e. trained on our own data) and our method trained on their data
significantly outperform their results, even at a 
stricter precision threshold. Sample results for this dataset are shown in
Figure \ref{fig:zhou-pies}A; note the regularity/simplicity of these charts,
compared to those in Figure \ref{fig:sample-results}.

Finally, to demonstrate the ease and success of extending our method to different chart types,
we trained our unmodified model on synthetic pie charts (dataset \ref{subsec:synthpie}). No change was made to our method;
only the annotations were replaced. Our method achieved an
F1 of 0.8319 (@1.5-cell thresh) on detection of salient pie chart elements from a holdout test set;
sample results are shown in Figure \ref{fig:zhou-pies}B. Numerous additional sample results are
provided in the supplemental material for all models and datasets.

\section{Discussion}
\label{sec:discussion}

This work shows that \textbf{a point-based method can very effectively detect data values in complex scientific bar charts with widely varying appearance}, and that synthetic data and simple augmentations can be well-leveraged to train such a model with efficient use of manually labeled data. Our approach is efficient (only 3\% larger than a ResNet50 model), easily extensible (as we demonstrate with pie charts) and designed to be readily incorporated in a data extraction pipeline or as a component of an end-to-end model.

\subsection{Extensibility}
\label{subsec:extensibility}
A key advantage we see in taking a point-based detection approach is the ready extensibility of this method to additional chart elements and chart types. One obvious example within the scope of bar charts is adding annotations for error bars as an additional point class and training the same model to predict these without modification. The same could be done with stacked charts, wherein multiple points could be labeled for each bar. In either case, the same model definition could be used, with the only change needed in the number of point classes. As demonstrated in Sec. \ref{subsec:lossfunc}, custom loss terms could also be added if desired and used to improve detection performance and consistency with particular chart expectations. Even very simple synthetic data (e.g. dataset \ref{subsec:synthdata}) can be used for model pre-training and augmented as discussed in Sec. \ref{subsec:realaug}.

For other chart types, the same approach holds. Histograms may be annotated and detected just like bar charts, with the only major differences being that the labeling of the data axis may be optional and the density of the data points may be higher (histogram bins may also be labeled differently than bars, but the alignment of such labels is independent of the detected data points). Scatter plots, dot plots and other point-based charts have a direct correspondence to a point-detection approach. We demonstrated that pie charts can be well described by labeling points at their center and the division between each wedge, and that our method can learn these without modification. Even line graphs can be detected as a dense set of connected points. In each case, salient data points can be detected with the same Point Proposal Network, with only annotation changes needed (and optionally scaling up for higher densities).

\subsection{Impact}
\label{subsec:impact}
The aim of this work is to extend researchers' ability to draw on information that is at present inaccessible to machine processing. We believe this effort and others like it will not only improve access to scientific knowledge, but also enable new uses of that knowledge, particularly when applied at large scales. Just as large-scale text mining has revolutionized natural language processing, so too do we think mining of charts and graphics can change the way the scientific literature is used as a resource for new research and discoveries. We also believe this type of work is aligned with broader efforts to improve scientific information access and usage (e.g. \cite{wilkinson2016fair}), particularly concerning the reusability of data. Outside the scientific literature, the method we propose can of course be applied and have value for examination and analysis of any source of charts (e.g. business, educational, government, etc.).

One important risk worth noting is the potential for over-reliance on ML-extracted information. As with other ML-based data mining techniques, the value detection method we propose is hopefully useful, but necessarily imperfect. Caution must of course be employed when using data extracted with such techniques, especially when done at scales that restrict human oversight. This is less of a concern if the method is used in an interactive setting.

\subsection{Limitations}
\label{subsec:limits}
This work does not address the full problem of bar chart value extraction from documents, but rather focuses on salient data point detection. As such, practical end-user utilization of the method we present would require two additional components: detection/cropping of charts in document images, and label detection/recognition (e.g. via OCR techniques). As we showed in Sec. \ref{sec:detectingcharts}, chart detection can be readily achieved with reasonably high accuracy using commonplace object detection methods such as YOLOv3. Customization of such models (e.g. to improve bounding box refinement) will certainly yield even better performance, though we developed our PPN value detection method and datasets with the expectation that some imperfection should be anticipated in chart crops.

Concerning label recognition, we designed our point-based approach such that its results can be incorporated with those of a text-detection model to complete the value extraction task. The position of the detected bar peaks and value-axis tick marks can be aligned with detected text to associate corresponding text labels to detected data points. We expect this approach will be particularly useful for detecting and extracting grouped bar charts, and for extracting values with nonlinear (e.g. logarithmic) or broken/segmented data axes as each tick mark can be labeled individually. Alternatively, our method can be combined with a chart detector and text label detector in a single model trained end-to-end, thus enabling direct chart value extraction from full document images (or collections of them), without the need for a pipeline approach.

Our decision to use a 56x56 feature map may prove a limitation for cases of particularly large or dense charts, though the method may of course be scaled up. The selection and generation criteria we used for our datasets -- particularly the exclusion of horizontal bar charts and omission of additional data points such as error bars -- is also not fully representative of the charts found in the scientific literature. Again, however, we anticipate that the point-based approach we pursued is particularly well suited to extending this method to these and other use cases.

\subsection{Future work}
\label{subsec:future}
We believe this work presents several opportunities for further research. Beyond extension to data extraction from other chart types (Sec. \ref{subsec:extensibility}), we've identified two particular research directions that merit investigation. The first is to combine our value detection method with a chart detector and text label detector in a single model, thus enabling end-to-end value extraction directly from full document images (or collections of them), without the need for a pipeline approach. The advantages of such an approach include reduction of performance bottlenecks, as well as access to additional information which may inform knowledge extraction (e.g. figure captions).

The second promising research avenue is the incorporation of an attention mechanism in the value extraction data path. If used to attend over a chart feature map, this could have particular value for identifying data groups, better predicting the number of data points, and recognizing spurious outliers. And if built into an end-to-end model as described above, attention could be further used to refine connections between multiple information classes (e.g. for knowledge graph generation informed by text, tables, and graphics).

\bibliographystyle{splncs04}
\bibliography{bib}
\end{document}